\documentclass[10pt,twocolumn,letterpaper]{article}
\usepackage[pagenumbers]{cvpr}
\usepackage[dvipsnames]{xcolor}
\definecolor{cvprblue}{rgb}{0.21,0.49,0.74}
\usepackage[pagebackref,breaklinks,colorlinks,citecolor=cvprblue]{hyperref}
\usepackage{xcolor}
\usepackage[normalem]{ulem}

\title{LegacyAvatars: Volumetric Face Avatars For Traditional Graphics Pipelines}

\author{
\begin{tabular}{c}
Safa C. Medin\textsuperscript{1,2} \qquad Gengyan Li\textsuperscript{1,3} \qquad Ziqian Bai\textsuperscript{1} \qquad Ruofei Du\textsuperscript{1} \qquad Leonhard Helminger\textsuperscript{1} \\[0.2em]
Yinda Zhang\textsuperscript{1} \qquad Stephan J. Garbin\textsuperscript{1} \qquad Philip L. Davidson\textsuperscript{1} \qquad Gregory W. Wornell\textsuperscript{2}  \\[0.2em]
Thabo Beeler\textsuperscript{1} \qquad Abhimitra Meka\textsuperscript{1} \\[0.8em]
\textsuperscript{1}Google \qquad \textsuperscript{2}MIT \qquad \textsuperscript{3}ETH Zurich 
\end{tabular}
\vspace{-1.2em}
}

\newcommand{\cM}{\mathcal{M}}   
   
\newcommand{\cT}{\mathcal{T}}   
   
\newcommand{\cW}{\mathcal{W}}   
\newcommand{\R}{\mathbb{R}}   

\begin{document}

\twocolumn[{
\renewcommand\twocolumn[1][]{#1}%
\maketitle
\begin{center}
  \newcommand{\teaserwidth}{\textwidth}
  \centerline{\includegraphics[width=\textwidth]{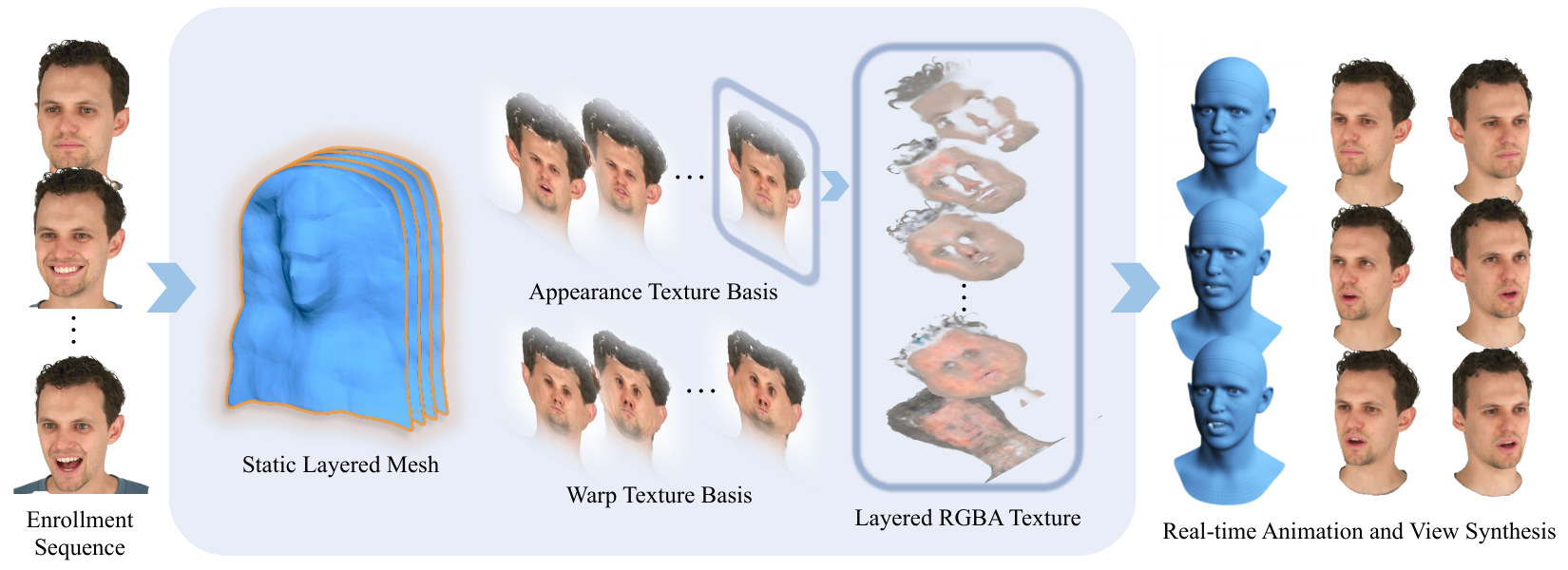}}
    \captionof{figure}{We present a novel representation for rendering animatable volumetric 3D face avatars using meshes and textures. From an enrollment sequence of a subject, we learn a layered mesh and blend-textures that model the geometry, appearance and deformations, and a simple linear transformation that maps tracked face model parameters to blend weights. Our representation is readily compatible with existing streaming infrastructure and can be deployed in traditional graphics pipelines in a device- and platform-agnostic way.}
    \label{fig:teaser}
\end{center}}]

\maketitle
\begin{abstract}
\vspace{-0.1cm}
We introduce a novel representation for efficient classical rendering of photorealistic 3D face avatars. 
Leveraging recent advances in radiance fields anchored to parametric face models, our approach achieves controllable volumetric rendering of complex facial features, including hair, skin, and eyes.
At enrollment time, we learn a set of radiance manifolds in 3D space to extract an explicit layered mesh, along with appearance and warp textures.
During deployment, this allows us to control and animate the face through simple linear blending and alpha compositing of textures over a static mesh. 
This explicit representation also enables the generated avatar to be efficiently streamed online and then rendered using classical mesh and shader-based rendering on legacy graphics platforms, eliminating the need for any custom engineering or integration. 
\footnotesize{
\url{https://syntec-research.github.io/LegacyAvatars/}}
\end{abstract}    
\section{Introduction}
\label{sec:intro}

Practical realization of real-time 3D face avatars~\cite{chen_authentic, Orts-Escolano2016Holoportation} demands progress on several fronts---capture (or enrollment), streaming, animation, and 3D rendering (or view synthesis). Over the last decade, modeling and rendering of these avatars have seen significant progress by enhancing surface-based geometry representation of parametric face models~\cite{egger20203d, flame} with volumetric components such as point clouds~\cite{Zheng2023pointavatar}, neural radiance fields~\cite{bai2023learningmonoavatar,hong2021headnerf,athar2022rignerf, insta}, or 3D Gaussians~\cite{ma2024gaussianblendshapes,qian2024gaussianavatars,GHA,chen2023monogaussianavatar}. Such volumetric techniques have improved the overall visual quality of the avatars and enabled data-driven modeling of the rigid and non-rigid dynamics of the human face~\cite{giebenhain2024npga, li2025tega, kirschstein2025avat3r, teotia2024gaussianheads}.

Current volumetric avatars still face several practical drawbacks that limit their large-scale, real-world applications.
\textit{First}, these representations are not natively compatible with the widely available rendering infrastructure. Most avatar applications on real devices are built using legacy platforms such as game engines (like Unity or Unreal) and design and rendering software (like Blender or Maya), which have been optimized over decades to efficiently use meshes and textures.
\textit{Second}, they are not immediately compatible with the existing streaming infrastructure for real-time applications, unlike 2D image and video streaming, which are widely used across devices and applications due to dedicated hardware, driver software, and compressed data formats. 
Modern volumetric representations typically parameterize the scene density and radiance with a neural network~\cite{nerf}, hash grids~\cite{mueller2022instant}, or 3D Gaussians~\cite{kerbl3Dgaussians}, 
which have been demonstrated with custom-built proprietary back-end infrastructure, but cannot easily operate across graphics platforms,  limiting their practical adoption~\cite{chen2022mobilenerf, tang2022nerf2mesh}.
Recently, game engine plugins have emerged that support static and dynamic volumetric scenes using 3D and 4D Gaussians~\cite{unity3dgs, lumaunreal}, 
but streaming \textit{animatable} or \textit{controllable} volumetric content like face avatars into such legacy graphics platforms remains an unsolved problem.

Motivated by these limitations, we develop a representation that allows \textit{animatable} volumetric rendering of face avatars using only legacy primitives like meshes and textures without relying on ML inference, facilitating native compatibility with standard graphics platforms. 
To achieve this, our key insight is to
discretize and quantize all continuous scene components of a face avatar---\textit{geometry}, \textit{appearance}, and \textit{deformation}---into classical primitives and modulate them using an animation control signal.
We draw inspiration from \textit{radiance manifolds (RM)} \cite{deng2022gram} that represent the volumetric radiance domain using a set of continuous non-intersecting surfaces. Previous works have leveraged RM for 3D playback of pre-recorded dynamic sequences~\cite{medin2024facefolds} by discretizing the surfaces and exporting them as a layered mesh and transparent RGBA texture that can be rendered through the standard graphics pipeline with additional alpha compositing. A similar paradigm was used for creating animatable avatars~\cite{bakedavatar} by learning deformable RM using a 3D morphable face model, which relies on neural networks to generate the animated appearance, leading to a complex rendering pipeline that disallows native compatibility and easy interoperability between graphics platforms.
While these methods utilize RM for discretizing scene geometry and appearance, they do not provide any means to discretize or explicitly control the scene \textit{deformation}.

We build on these ideas by tackling the problem of discretizing and controlling the \textit{deformation} and \textit{animation}. {We uniquely show that animation control does not require deforming the layered mesh vertices, but simply offsetting the UV coordinates that are used for sampling the RGB and alpha textures. We also show that these UV-offsets for a given expression can be modeled using a simple linear transform of the tracked face model parameters.}
Our resulting assets consist of a single layered mesh and a basis of appearance and UV-warp texture maps that need to be streamed \textit{only once} at the beginning, using standard mesh and image compression. We achieve temporal animation of the avatar by streaming only the tracked face model parameters (\(\sim \)2KBs per frame), which are then linearly transformed to linear-blend coefficients for the texture bases to achieve the final layered appearance texture. Our assets can be rendered on any device using a single pass of a simple programmable shader for linear compositing followed by standard rasterization.
In summary,
\begin{itemize}
    \item We present a novel representation for volumetric 3D face avatars that models geometry, appearance, \textit{and} deformation of the scene using only legacy graphics primitives of a layered triangle mesh and a set of textures, \textit{without} relying on ML inference for rendering. This allows native compatibility with widely available graphics platforms and online streaming infrastructure.
    \item Through qualitative and quantitative comparisons on a publicly available dataset~\cite{wuu2022multiface}, we demonstrate that our representation achieves efficient rendering on a traditional graphics platform like WebGL on a consumer laptop, while maintaining volumetric visual quality.
\end{itemize}

\section{Related Work}
\label{sec:related-work}

\noindent\textbf{Early methods.} The first real-time 3D avatar systems were realized using differentiable 3D morphable face models (3DMMs)~\cite{3dmm,egger20203d}, enabling efficient optimization frameworks for canonical performance capture and playback. 
But these models typically suffer from insufficient representational capacity since they are inherently low-dimensional and cannot model complex, high-frequency effects in geometry and appearance, limiting the output visual quality. 

\begin{figure*}[t]
  \newcommand{\teaserwidth}{\textwidth}
  \centerline{\includegraphics[width=\textwidth]{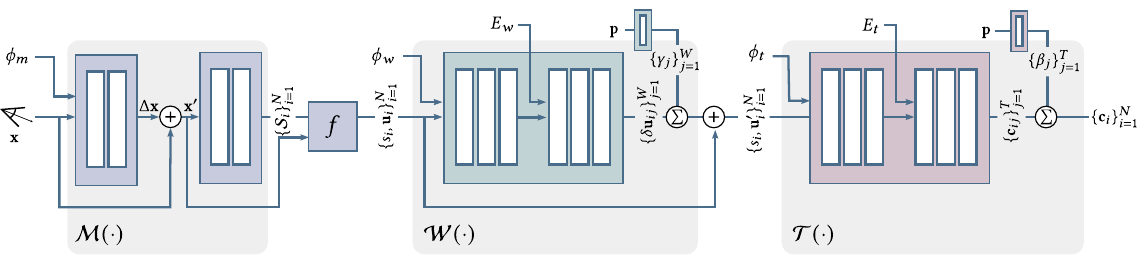}}
    \captionof{figure}{\textbf{Training pipeline for enrollment phase.} Our model consists of three separate modules: a manifold predictor $\cM$, a warp predictor $\cW$, and a texture predictor $\cT$. Here, $\cM$ is a scalar field that defines layered implicit surfaces. The intersections with these surfaces are spherically mapped to the UV-space via a learnable function $f$. Then, the output subsequently queries $\cW$ to obtain a basis of UV-offsets. These offsets are then linearly blended as a function of expression parameters of a face model and added to the original values. Finally, the new coordinates are fed through $\cT$, which predicts a basis of RGBA appearances that are also linearly blended as a function of expression parameters. Each module takes in learned latent codes $\phi_m, \phi_w, \phi_t$ for multi-subject training, while $\cW$ and $\cT$ take in learnable embedding matrices $E_w$ and $E_t$ to output bases of warps and textures.} 
    \label{fig:pipeline}
\end{figure*}
 
\vspace{0.2cm}\noindent\textbf{Volumetric methods.} Complex geometry and appearance of faces, including hair and eyes, have recently been modeled using volumetric methods. \citet{park2021nerfies,park2021hypernerf} show that radiance fields can be combined with a learned deformation field to reconstruct a high-quality 3D canonical face model, including minute details such as hair strands. Such ideas have been extended to real-time avatars by enhancing 3DMMs with volumetric radiance fields~\cite{hong2021headnerf,athar2022rignerf,Gafni_2021_CVPR,chen_authentic,bai2024efficient,liu2021neuralactor,shellnerf}. More recently, point clouds \cite{Zheng2023pointavatar} and 3D Gaussians~\cite{qian2024gaussianavatars,GHA, saito2023rgca,chen2023monogaussianavatar,ma2024gaussianblendshapes, wang2024mega,li2025tega} have been used to achieve best-in-class real-time avatars using similar strategies, but
they suffer from other challenges that limit their applications. For example, Gaussian splatting (GS) implementations are typically fine-tuned for specific compute architectures like GPUs, hence cannot be easily deployed in a hardware-agnostic setting. GS also suffers from the famous issue of \textit{popping} phenomenon due to the primitive ordering during view-dependent rendering, causing distracting artifacts in the scene. While there exist methods that introduce sorting-free GS~\cite{hou2024sort,radl2024stopthepop}, they are tailored for general, static scenes. Finally, Gaussian primitives are not natively compatible with the existing streaming infrastructure, hence streaming them requires specialized engineering~\cite{sun20243dgstream}.

Several avatar representations have focused on efficient engineering by discretizing the 3D volume to achieve real-time results. Tri-plane representations learn features in orthogonal planes and project them to 3D~\cite{gnarf,next3d,3dpr,lite2relight}. Hash grids~\cite{insta} and hash ensembles~\cite{kirschstein2023nersemble,bai2024efficient} voxelize the 3D space while promoting quick spatial queries, leading to real-time rendering of radiance fields. Tetrahedral fields have been used to directly deform a volumetric representation to realize animatable avatars~\cite{garbin2022voltemorph, yuan2022nerfediting, kania2023blendfields}. {But similar to Gaussian-based approaches, these methods also require custom engineering for deployment to real-world scenarios.}

Another class of methods extends classical mesh-based blendshapes to volumetric representations by introducing blendable bases using neural fields~\cite{nerfblendshape} or Gaussian primitives~\cite{ma2024gaussianblendshapes}. Although our method follows a similar principle---as we also combine a fixed set of assets to synthesize novel expressions---these methods rely on explicit 3D deformations to represent changes in the face geometry. Moreover, like their conventional neural field or 3D Gaussian-based counterparts, they are not immediately suitable for practical telepresence systems.

\vspace{0.2cm}\noindent\textbf{Implicit surface-based methods.} We draw inspiration from a class of methods that learn a collection of implicit surfaces~\cite{deng2022gram, Esposito2025VolSurfs} to discretize the 3D volume~\cite{yue2022anifacegan,bakedavatar,medin2024facefolds,Xu2023:LayeredSurfaceVolumes}. These methods were first used in a GAN-based pipeline to generate 3D renderings of novel human identities \cite{deng2022gram} and then extended to control their generated pose or expression \cite{Xu2023:LayeredSurfaceVolumes,yue2022anifacegan}. FaceFolds~\cite{medin2024facefolds} uses radiance manifolds to model pre-recorded videos of dynamic face performances of real people and exports them as a static layered mesh textured with an RGBA video, demonstrating compatibility with traditional 3D rendering and streaming systems. But it does not explicitly discretize or control face deformations and merely captures them in appearance textures for playback, providing no means to control or animate the faces. We build on these ideas to achieve animation by 1) introducing a warping mechanism to represent most geometric changes in UV-space deformations instead of offloading all variations to alpha composition of appearance textures, 2) representing warp and appearance as linear combinations of a fixed basis of assets to achieve ML-free rendering, and 3) leveraging synthetic data to enable generalization in the low-dimensional expression parameter space without realizing the full 3DMM mesh.

An avatar system most similar to our representation is BakedAvatar~\cite{bakedavatar}, which similarly extracts layered mesh and textures from learned implicit surfaces. But differently, they perform face animation by explicitly deforming their layered meshes using per-vertex FLAME deformation weights~\cite{flame}. 
In the fragment shader, they rely on an MLP to dynamically compute blend weights, which linearly combine multiple pre-baked textures to produce the final pixel color.
Since these non-linear deformations and neural network inference demands custom engineering during deployment, their representation is not natively compatible with legacy renderers. In addition, their per-pixel MLP queries also make scaling to higher resolutions increasingly more expensive.  Whereas in our method, we 1) model deformations discretely and linearly in the UV space of a static mesh without having to deform it, and 2) do not rely on any neural networks at deployment time, which helps us achieve real-time performance at high resolutions on consumer devices, as well as backward compatibility and streamability trivially.
We perform a thorough evaluation against BakedAvatar~\cite{bakedavatar} to adequately demonstrate, both qualitatively and quantitatively, the improvements our method offers---in terms of both visual quality \textit{and} compute efficiency and while,  most importantly, being readily compatible with legacy graphics pipelines.
\section{Methodology}
\label{sec:method}

A 3D avatar system generally consists of two phases, an \textit{enrollment} phase in which the avatar is created using data captured from a subject, and a \textit{deployment} phase in which the avatar is streamed, animated, and rendered on the client device from a desired viewpoint. Our goal is to design a volumetric avatar representation that is compatible with legacy rendering platforms during the deployment phase. We achieve this by exporting our enrolled volumetric avatar to classical graphics primitives like meshes and textures that can be rendered efficiently using simple programmable shaders on any graphics platform without additional custom engineering, agnostic of the underlying device hardware or software. We also aim to make our representation conducive to online streaming, which includes the ability to trade off quality and data bandwidth via data compression, similar to today's online video streaming systems.

\subsection{Avatar as Layered Mesh} 
Given the calibration video of a subject, our objective is to learn a single layered mesh representation of the subject that can be dynamically textured using expression coefficients of a parametric face model in real-time. {Similar to previous methods, we use radiance manifolds \cite{deng2022gram} to model the geometry as a set of static 2D implicit surfaces \cite{medin2024facefolds,Xu2023:LayeredSurfaceVolumes}, and the appearance as a UV-mapped dynamic radiance controlled by the 3DMM coefficients \cite{bakedavatar}}.

Our geometry is modeled by learning a set of $N$ implicit surfaces $\{ \mathcal{S}_i\}_{i=1}^N$ defined by a single manifold predictor $\cM: \R^3 \to  \R$, which takes in a 3D point $ \mathbf{x} \in \R^3$ and outputs a scalar value $s \in \R$. We first map all points on these surfaces via a \textit{learnable} function $f: \mathcal{S}_i \to [-1,1]^2$ to obtain UV-space coordinates. In the UV-space, we learn a set of $W$ warp fields and $T$ RGBA texture fields for each surface by parameterizing the following functions as MLPs: 
\begin{align}
    \cW_{ij} &:  \mathbf{u} \mapsto \delta \mathbf{u} \\
    \cT_{ik} &: \mathbf{u} \mapsto \mathbf{c}\,,
\end{align}
where $\mathbf{u} \in [-1,1]^2$ is a UV-space coordinate, $\delta \mathbf{u} \in \R^2$ is a UV-space offset, $\mathbf{c} \in \R^d$ includes a scalar alpha value and view-dependent color parameterized using spherical harmonics~\cite{ramamoorthi2001efficient}, and $i \in \{1, 2, \dots, N\}$, $j \in \{1, 2, \dots, W\}$, $ k \in \{1, 2, \dots, T\}$.
Given a set of face model parameters $\mathbf{p} \in \R^p$ of a single frame, a layered warp field $\{\cW_{i}\}_{i=1}^N$ and a layered texture field $\{\cT_{i}\}_{i=1}^N$ are obtained by
\begin{equation}
    \cW_{i} = \sum_{j=1}^W \gamma_j \cW_{ij} \quad \text{and} \quad \cT_{i} = \sum_{k=1}^T \beta_k \cT_{ik} 
\end{equation}
where $\{\gamma_j\}_{j=1}^W$ and $\{\beta_k\}_{k=1}^T$ are a set of weights obtained as a learnable \textit{linear} function of $\mathbf{p}$.  

Given a 3D point $\mathbf{x} \in \mathcal{S}_i$ and its UV-coordinate $\mathbf{u}$, we compute the warped UV-values $\mathbf{u}' = \mathbf{u}+ \cW_i(\mathbf{u})$, which are used to query the blended texture field to obtain color and transparency as $\mathbf{c} = \cT_i(\mathbf{u}')$. We design our pipeline in a way that the implicit surfaces and the warp and texture bases can be efficiently discretized and exported into a layered mesh and UV-space maps in pixel-space, allowing them to be immediately deployed to any graphics platform without relying on custom rendering algorithms or ML tools.

\subsection{Dataset}
\label{sec:dataset}
To train our model, we use multiview videos from the NeRSemble dataset~\cite{kirschstein2023nersemble}, which consists of a set of subjects with various facial expressions and talking sequences. We fit a parametric face model to the video sequences of each subject to obtain per-frame expression coefficients as well as per-pixel UV-values to explicitly supervise correspondences between different frames. As a preprocessing step, we transform the camera extrinsics to align faces across frames to a canonical 3D face in order to account for strong head rotations during the capture. We downsample the original images to $802\times550$ resolution.

\vspace{0.2cm}\noindent\textbf{Expression generalization.}
Learning an expressive geometry and appearance model that also generalizes to novel expressions outside of the calibration sequence is a challenging and ill-posed problem. Furthermore, sampling across 2D discrete surfaces instead of the entire 3D volume introduces more instability in training, which can cause shelling artifacts in extreme poses~\cite{medin2024facefolds}. To aid generalization to novel expressions and to mitigate stability issues arising in joint learning of geometry and appearance, we introduce a multi-subject training paradigm. Uniquely, we make use of a publicly available synthetic multi-view multi-expression image dataset~\cite{buehler2024cafca}. In our experiments, we combine each subject in the NeRSemble dataset~\cite{kirschstein2023nersemble} with a number of synthetic subjects that have similar face shapes to the real subject. We show that such joint training helps avoid overfitting to the expressions from the calibration sequence of the target subject and prevents artifacts. Please see the supplementary results for more details.

\subsection{Model Architecture and Training}
Our architecture consists of three modules: a manifold predictor $\cM$, a UV-space warp predictor $\cW$, and a texture predictor $\cT$, which we illustrate in~\cref{fig:pipeline}.

\vspace{0.2cm}\noindent\textbf{Manifold predictor.} 
We implement our manifold predictor as a \textit{subject-specific} scalar field that takes in 3D coordinates in $xyz$-space and a learned latent code $\phi_m$ for each subject. This module defines a set of $N$ implicit surfaces, which are \textit{static} for the entire sequence of a given subject. To achieve this, an input 3D point is first deformed using a subject-specific deformation field, and subsequently used to predict a scalar value that determines the manifold geometry. Given a camera ray, we first find $xyz$-space intersections of this ray with each manifold~\cite{niemeyer2020differentiable} and compute an initial set of UV-coordinates using spherical mapping~\cite{medin2024facefolds}. But since we are interested in learning dense UV-space correspondences across frames consistent with the ground truth UV values, this spherical transformation is done with respect to a scene center set as a 3D learnable parameter instead of a fixed one.

\vspace{0.2cm}\noindent\textbf{UV-space warp predictor.} Uniquely in our animation model, we learn a basis of UV-space offsets for each of the surfaces implemented as an MLP, which receives 
a learned subject code $\phi_w$ and a set of $W$ learned latent codes represented as an embedding matrix $E_w$. 
This defines a set of warp fields that can be linearly combined into a single warp field, which is used to add offsets to the input UV-coordinates before querying the appearance model. In particular, we first \textit{linearly} map the per-frame expression coefficients to a set of weights that \textit{linearly} blend the predicted UV warps across different surfaces. We note that the blending weights are the same for all surfaces, facilitating a lightweight compute at rendering time.

\begin{figure}[t]
  \centering
  \includegraphics[width=0.95\linewidth]{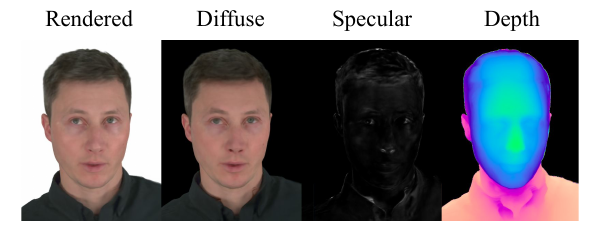}
  \caption{\textbf{Radiance decomposition.} During training, the radiance is modeled using spherical harmonics coefficients, which can be decomposed into diffuse (view-independent) and specular (view-dependent) components. The appearance can be exported as just diffuse or both diffuse and specular texture images.
  \vspace{-0.3cm}}
  \label{fig:diff-spec}
\end{figure}

\vspace{0.2cm}\noindent\textbf{UV-space texture predictor.}
To account for the geometry and appearance changes that cannot be fully modeled by the UV-offsets (such as the inner mouth, eyelid motions, or complex specularities on the skin), we learn a set of UV-space {explicit} blend-textures that can also be linearly combined into a single layered texture using the 3DMM expression parameters. Our texture predictor receives 
a learned subject code $\phi_t$ and a set of $T$ learned latent codes represented as an embedding matrix $E_t$,
and it outputs $2$-degree spherical harmonics coefficients for radiance~\cite{ramamoorthi2001efficient} and a scalar alpha value. 
Finally, given a UV-space coordinate $\mathbf{u}'$ and a view direction in $xyz$-space, the output of our entire pipeline is an RGB radiance and an alpha value, which are composited across rays to produce the final color:
\begin{equation}
    \mathbf{w}_i \triangleq \alpha_i \prod_{j=1}^{i-1} (1-\alpha_j) \quad \mathbf{c}= \sum_{i=1}^N \mathbf{w}_i \mathbf{c_i}
\end{equation}
At inference time, we decompose the radiance into diffuse and specular components, which provides further flexibility on the size of the assets that are exported from the model. In~\cref{fig:diff-spec}, we illustrate our renders for each component.

\begin{figure}[t]
  \centering
  \includegraphics[width=0.9\linewidth]{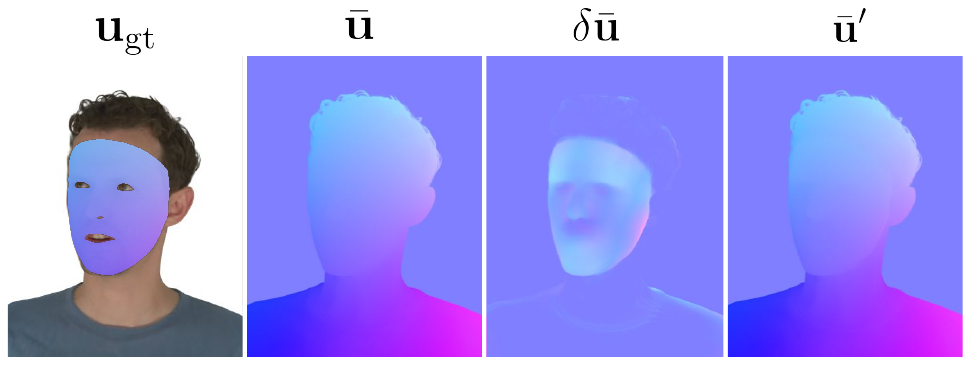}
  \caption{\textbf{UV-space predictions.} Given a ground truth per-pixel UV-coordinates $\mathbf{u}_\mathrm{gt}$, our model is supervised to match the \textit{expectation} of warped coordinates $\bar{\mathbf{u}}'$ to the ground truth. We visualize the expectations of the spherically mapped coordinates $\bar{\mathbf{u}}$ and the warps $\delta\bar{\mathbf{u}}$ for reference.  \vspace{-0.5cm}}
  
  \label{fig:uv-pred}
\end{figure}

\vspace{0.2cm}\noindent\textbf{Loss functions.}
We train our pipeline in an end-to-end fashion by adopting the following loss function:
 \begin{equation}
    \mathcal{L} = \mathcal{L}_\mathrm{rec} + \lambda_{\mathrm{uv}} \mathcal{L}_\mathrm{uv} + \lambda_{\mathrm{silh}} \mathcal{L}_\mathrm{silh}  +  \lambda_{\mathrm{reg}} \mathcal{L}_\mathrm{reg}
 \end{equation}
 where $\mathcal{L}_\mathrm{rec}$ is an $\ell_1$ reconstruction loss between predicted and ground truth pixel values, $\mathcal{L}_\mathrm{uv}$ is an $\ell_1$ loss between the per-pixel \textit{expected} UV-coordinates and ground truth UV-values computed only on skin regions of the face, $\mathcal{L}_\mathrm{silh}$ is a silhouette loss that guides the geometry of each layer using per-image foreground masks~\cite{bakedavatar}, and $\mathcal{L}_{\mathrm{reg}}$ is the regularization term that penalizes predicted warps, specular radiances, and manifold predictor weights to promote training stability~\cite{medin2024facefolds}.
 Here, the expected UV for each pixel is obtained by a weighted combination of the predicted warped UVs across rays $\{ \mathbf{u}'_i\}_{i=1}^N$ as $\bar{\mathbf{u}}' \triangleq \sum_{i=1}^N \mathbf{w}_i \mathbf{u}'_i$, which is supervised to match the ground truth UV-values at that pixel. The UV supervision is a key term to improve the model capacity by registering the facial features consistent with the UV topology of the 3DMM, so that the texture basis focuses on appearance changes that cannot be explained by UV-space warps. We illustrate a representative test frame along with its per-pixel UV values and warps in~\cref{fig:uv-pred}.

\vspace{0.2cm}\noindent\textbf{Training details.}
The manifold predictor is implemented as two cascaded $4$-layer MLPs of widths $128$, which respectively learn a subject-specific 3D warp field and a scalar field that determines the manifold geometry. The warp predictor is a $6$-layer MLP of widths $128$, which takes in $W$ learnable embeddings of dimension $128$ after the third layer. The texture predictor is a $6$-layer MLP of widths $256$, which receives $T$ learnable embeddings of dimension $128$ after the third layer. The subject embeddings are $128$-dimensional vectors, which condition each module individually. We optimize our entire model using the Adam optimizer~\cite{adam} with initial learning rates of $0.0007$, $0.0005$, $0.0008$ and exponential decay rates of $0.20$ per $200\,000$ iterations for $\cM$, $\cW$, and $\cT$, respectively. With a batch size of $32\,768$ rays sampled across all subjects, frames, and views, we train our pipeline for $500\,000$ iterations, 
which takes approximately 36 hours over $8$ NVIDIA H100 GPUs.

\subsection{Exporting 3D Assets and Model Deployment}
Similar to previous works~\cite{medin2024facefolds}, we discretize our continuous manifolds by shooting rays from a hemisphere (centered at the learned scene center) uniformly in azimuth and elevation angles, gathering all intersections and topologizing them into a triangle mesh. We use the same set of points to query our warp and texture predictors to obtain a basis of UV-space offsets and appearance maps. {Uniquely, we export the linear functions that map expression coefficients to blend weights as individual matrices, allowing us to control animation without deforming the mesh but through simple blending of our warp textures}. Depending on the application, the mesh can be decimated to reduce the number of primitives, while UV-offset and appearance maps can be downsampled to lower resolutions. Our final assets merely comprise a single layered mesh with a fixed topology as well as warp and appearance maps, and they can easily be deployed to any graphics platform for rendering.

\subsection{Rendering}

\begin{figure}[t]
  \centering
  \includegraphics[width=1.0\linewidth]{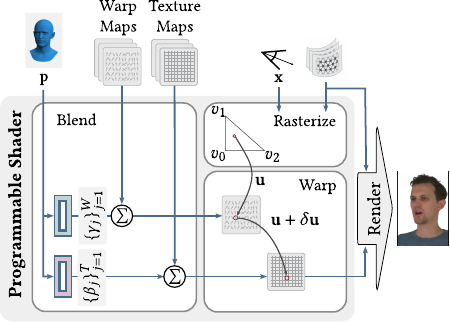}
  \caption{\textbf{Programmable shader.} At the deployment phase, our 3D assets (a single static layered mesh and bases of warp and texture maps) can easily be used to render dynamic and volumetric faces via a programmable shader on any graphics platform.
  \vspace{-0.5cm}}
  \label{fig:prog-shader}
\end{figure}

The linear blending and alpha compositing is performed by a programmable shader that receives the exported 3D assets, as well as face model parameters $\mathbf{p}$ and the camera viewpoint, as shown in Fig.~\ref{fig:prog-shader}. Here, blend and warp phases are simple linear operations, while rasterization is a projection operation handled by the {standard} graphics pipeline. By learning a canonical geometry and modeling deformations in the 2D UV-space instead of the 3D space, we can disentangle their parameterization into a single mesh and a set of warp maps. In our results, we show that it is possible to model the blend weights for the warp and appearance bases as a linear transformation of the expression parameters. This makes animation a very simple linear operation without having to explicitly account for complex non-rigid dynamics in 3D. This is in contrast to mesh-based avatar methods such as BakedAvatar~\cite{bakedavatar}, which handles animations by explicitly deforming meshes.

\vspace{0.2cm}\noindent\textbf{A note on rendering efficiency in comparison with modern methods.}
Rasterizing triangles is inherently significantly faster than the standard implementations of volumetric rendering techniques such as Gaussian splatting and neural fields on a per-primitive and per-pixel basis. Gaussian splatting requires an expensive sorting operation, while neural fields rely on tracing rays, sampling multiple points along the ray that need to be contiguously integrated. Rasterizing our ordered mesh representation does not suffer from such challenges since sorting is handled using the depth buffer.
Finally, we \textit{do} note that there are several implementations of 3D Gaussian splatting and neural fields, including hash grids like InstantNGP~\cite{mueller2022instant} and hierarchical embeddings~\cite{kerbl2024hierarchical} that offer faster results, but they are often engineered for particular hardware such as a GPU or require custom implementations for wide-scale deployment.

\subsection{Streaming}

Our method simplifies data transmission by initially sending the static mesh, blend textures, and linear transformations \textit{only once}. Subsequently, only per-frame face model parameters $\mathbf{p}$ are streamed, which are fed directly to the programmable shader in Fig.~\ref{fig:prog-shader}.  
Our technique also offers a unique advantage in \textit{multi-avatar interaction} scenarios. In a 1-on-1 interaction, complete client-side rendering is ideal, as it minimizes the amount of data that must be transmitted. But in a multi-avatar scenario, offloading compute to the server side is more efficient as it avoids duplication of compute effort across multiple clients. Our pipeline allows for expression-related computations such as warping and blending to be performed on the server side, so that only the view-dependent rendering is left to the client side. Most importantly, since the output of our warping and blending operations is a set of texture maps, they can be conveniently transmitted as a compressed video stream. This allows for a healthy trade-off between compute and transmission bandwidth.
Such a trade-off is not trivially available with other volumetric techniques, including other layed-mesh based techniques like BakedAvatar~\cite{bakedavatar}.
\section{Experiments and Results}

\begin{figure}[t]
  \centering
  \includegraphics[width=0.8\linewidth]{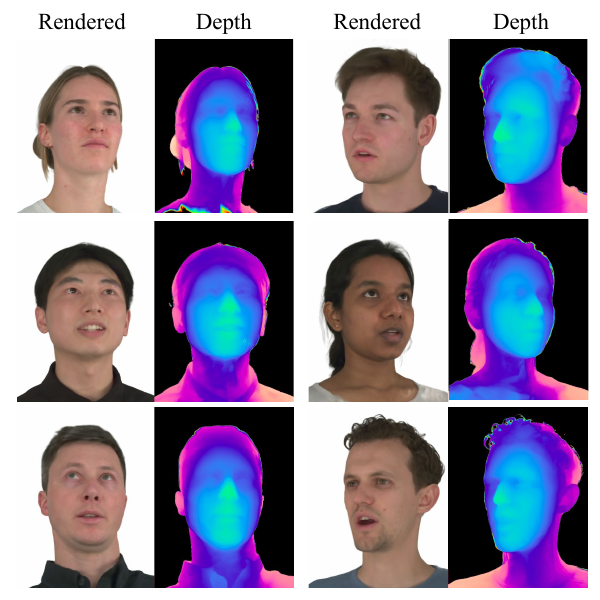}
  \caption{\textbf{Novel view synthesis results.} Our model achieves photorealistic volumetric rendering of 3D face avatars. Please see the supplementary material for video demonstrations.
  \vspace{-0.3cm}}
  \label{fig:qual-results}
\end{figure}

In our experiments, we set $N=T=W=12$. Following FaceFolds~\cite{medin2024facefolds}, we select a number of manifolds that allows for volumetric effects while maintaining rendering efficiency. Similarly, basis sizes are tuned to balance computational efficiency with the model’s expressivity.
We use the first $9$ talking sequences of each subject from the NeRSemble dataset~\cite{kirschstein2023nersemble} for training and use the last one for testing. We hold out $2$ of the $16$ cameras in training data to evaluate our model's performance quantitatively. To aid the training stability and generalization, we gather $25$ subjects from the synthetic data corpus with the closest head shapes to the real subjects, as discussed in Sec.~\ref{sec:dataset}. 
We measure this proximity using a variance-weighted Euclidean distance between the PCA identity vectors.

We compare against five state-of-the-art efficient volumetric avatar techniques, BakedAvatar~\cite{bakedavatar}, Gaussian Head Avatar (GHA)~\cite{GHA}, GaussianAvatars~\cite{qian2024gaussianavatars},  MonoAvatar++~\cite{bai2024efficient}, and PointAvatar~\cite{Zheng2023pointavatar}. We chose these methods as representative techniques that achieve fast rendering and employ layered meshes, Gaussian splatting, or neural radiance fields as the underlying volumetric representation. \textbf{Note:} The goal of our method is to achieve volumetric effects using legacy primitives and no ML inference. We do \textit{not} claim that we outperform continuous volumetric avatar techniques that are based on Gaussians or NeRFs on overall visual quality. We only show comparable results with a sample of such methods in order to visually place our technique in the overall context of the state-of-the-art.

\subsection{Qualitative Results}

\begin{figure}[t]
  \centering
  \includegraphics[width=1.0\linewidth]{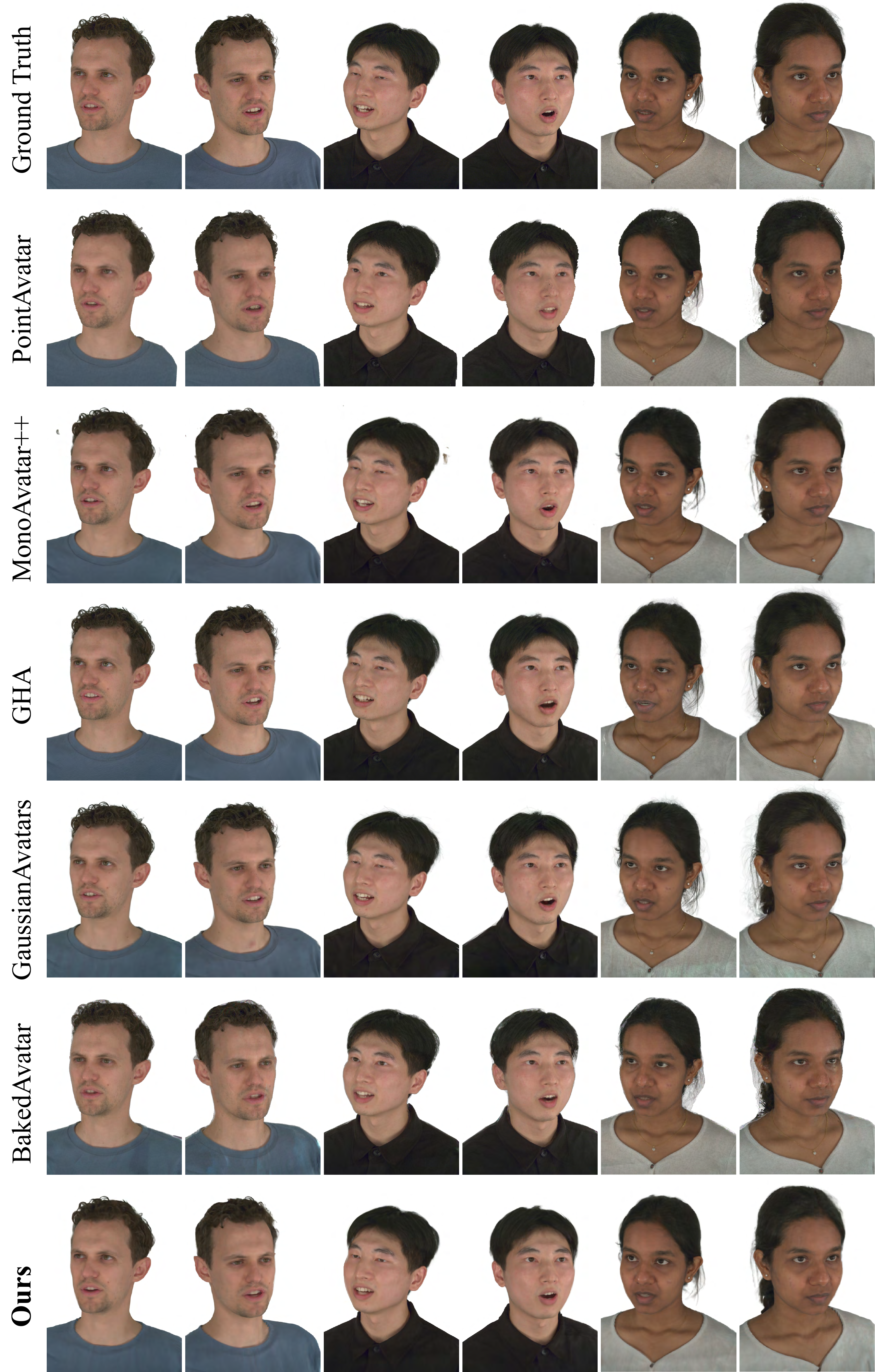}
  \caption{\textbf{Qualitative comparisons.} Our technique achieves comparable visual quality to modern neural
rendering techniques while facilitating 3D animations in a platform-agnostic way.
\vspace{-0.2cm}}
  \label{fig:qual-comp}
\end{figure}

\vspace{0.2cm}\noindent\textbf{Novel view synthesis.}
We illustrate our novel view synthesis results in~\cref{fig:qual-results}. Our method generalizes over different subjects with varying face geometries and appearances. Please refer to the supplementary material for video demonstrations and comparisons with other methods.

\vspace{0.2cm}\noindent\textbf{Self- and cross-reenactment.}
We show renders of test expressions from the held-out views and compare them with baseline methods in~\cref{fig:qual-comp}. Our model achieves comparable visual quality to the previous methods that rely on sophisticated primitives or MLP queries at rendering time. Please refer to the supp.\ video for cross-reenactment results.

\begin{table}
\footnotesize
  \caption{\textbf{Quantitative comparisons.} Our model attains similar performance compared to the state-of-the-art methods.}
  \label{table:comparisons}
  \begin{tabular}{cccc}
    \toprule
    Method & PSNR\,$\uparrow$ & SSIM\,$\uparrow$ & LPIPS\,$\downarrow$ \\
    \midrule
    PointAvatar& $23.80\!\pm\!1.28$ & $0.872\!\pm\!0.016$ & $0.137\!\pm\!0.018$\\
    MonoAvatar++ & $27.45\!\pm\!2.43$ & $0.936\!\pm\!0.011$ & $0.098\!\pm\!0.009$\\
    GHA & $24.29\!\pm\!2.16$ & $0.863\!\pm\!0.039$ & $0.102\!\pm\!0.024$\\
    GaussianAvatars & $27.54\!\pm\!1.69$ & $0.931\!\pm\!0.017$ & $0.066\!\pm\!0.015$\\
    BakedAvatar & $24.38\!\pm\!0.78$ & $0.888\!\pm\!0.013$ & $0.117\!\pm\!0.018$\\
    Ours & $26.97\!\pm\!1.23$ &  $0.929\!\pm\!0.007$ & $0.117\!\pm\!0.006$ \\
    \bottomrule
\end{tabular}
\end{table}

\subsection{Quantitative Results}
For three subjects (with IDs \texttt{055}, \texttt{264}, and  \texttt{306}), we perform quantitative evaluations on two held-out views across the entire test sequences, consisting of over $800$ images. We report average image quality metrics in PSNR, SSIM~\cite{ssim}, and LPIPS~\cite{lpips} for our method and other methods in~\cref{table:comparisons}, where we observe comparable average performance, with differences falling within the margin of variance.

\subsection{Ablation Studies}
\begin{figure}[t]
  \centering
  \includegraphics[width=1.0\linewidth]{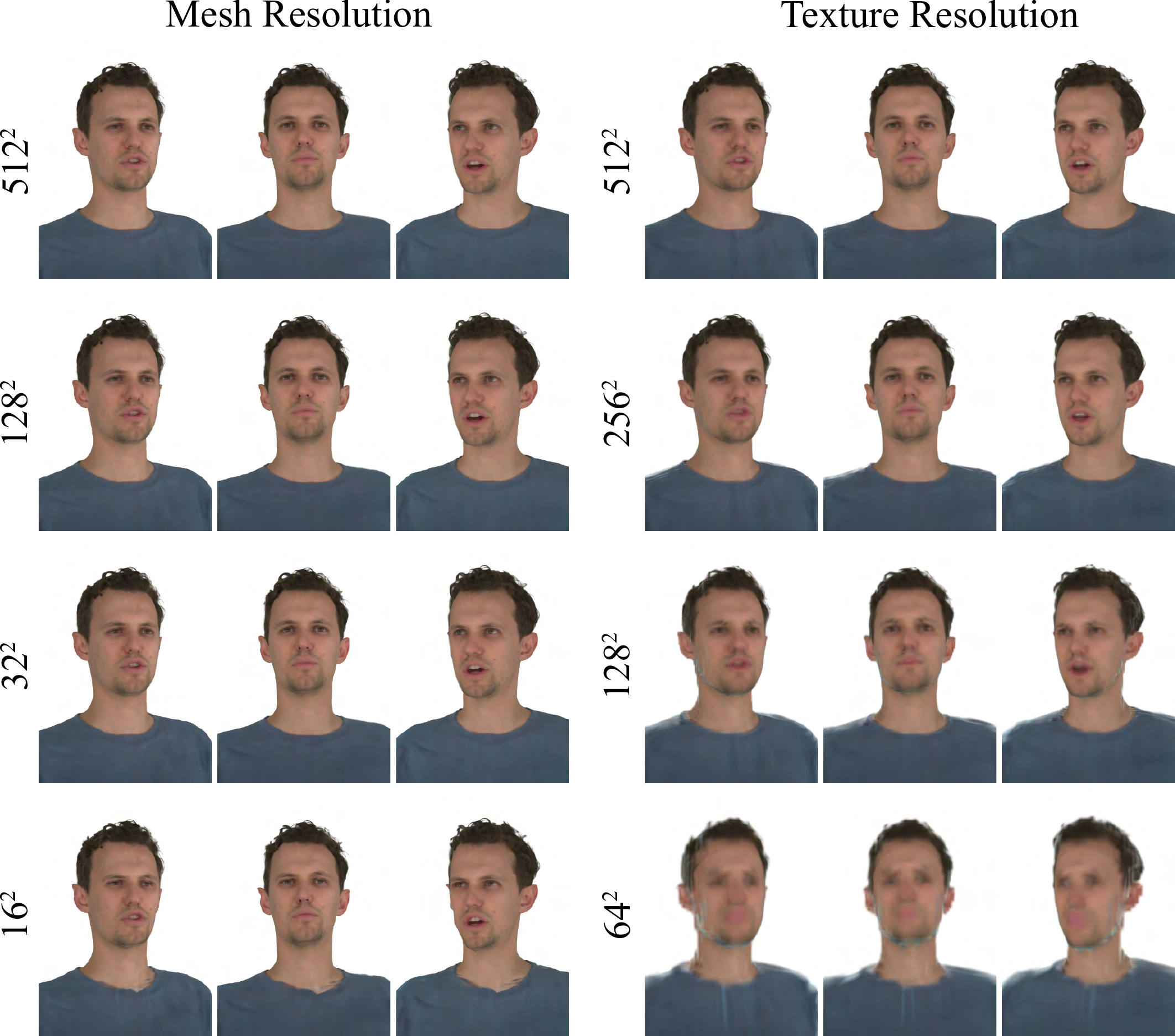} 
  \caption{\textbf{Ablation on mesh and texture resolution.} At the deployment phase, our asset size can be trivially reduced with standard operations. Due to our smooth surface geometry, the visual quality is maintained down to $32\times32$ mesh resolution, with a total number of primitives of $<\!\!12\,000$, providing a very lightweight volumetric representation for renders at $0.5$K resolution. Similarly, the texture resolution can also be adjusted for varying needs of an application by downsampling layered textures.
  \vspace{-0.5cm}}
  \label{fig:resolution-ablation}
\end{figure}

\paragraph{Mesh and texture resolution.}
Similar to FaceFolds~\cite{medin2024facefolds}, our representation has the flexibility to efficiently trade off image quality with rendering efficiency by reducing the number of primitives of the exported mesh and downsampling the layered textures. Given a layered mesh at $512\times512$ vertex resolution (per layer) with canonical texture coordinates as vertex attributes, we first gather the vertices from each layer as an oriented point cloud and perform Poisson surface reconstruction~\cite{kazhdan2006poisson} to obtain watertight surfaces~\cite{medin2024facefolds}. Then, we use an off-the-shelf mesh decimation algorithm to reduce the number of vertices in the mesh to a given target. Since our model is trained to represent a variety of expressions with a single set of static surfaces, the exported meshes roughly correspond to the coarse face geometry of the subjects. Therefore, we can reduce the total number of primitives significantly without sacrificing the visual quality, see \cref{fig:resolution-ablation} and the supplementary video.

If there is any need to decrease the resolution of the streamed avatar (such as reduced data bandwidths), we can dynamically downsample our video textures using the existing infrastructure. We report view synthesis and animation results for a variety of texture resolutions, illustrated in \cref{fig:resolution-ablation} and the supplementary video. For more ablations, please refer to the supplementary material.

\subsection{Real-time Rendering on Web Browsers}
\begin{figure}[t]
  \centering
  \includegraphics[width=1.0\linewidth]{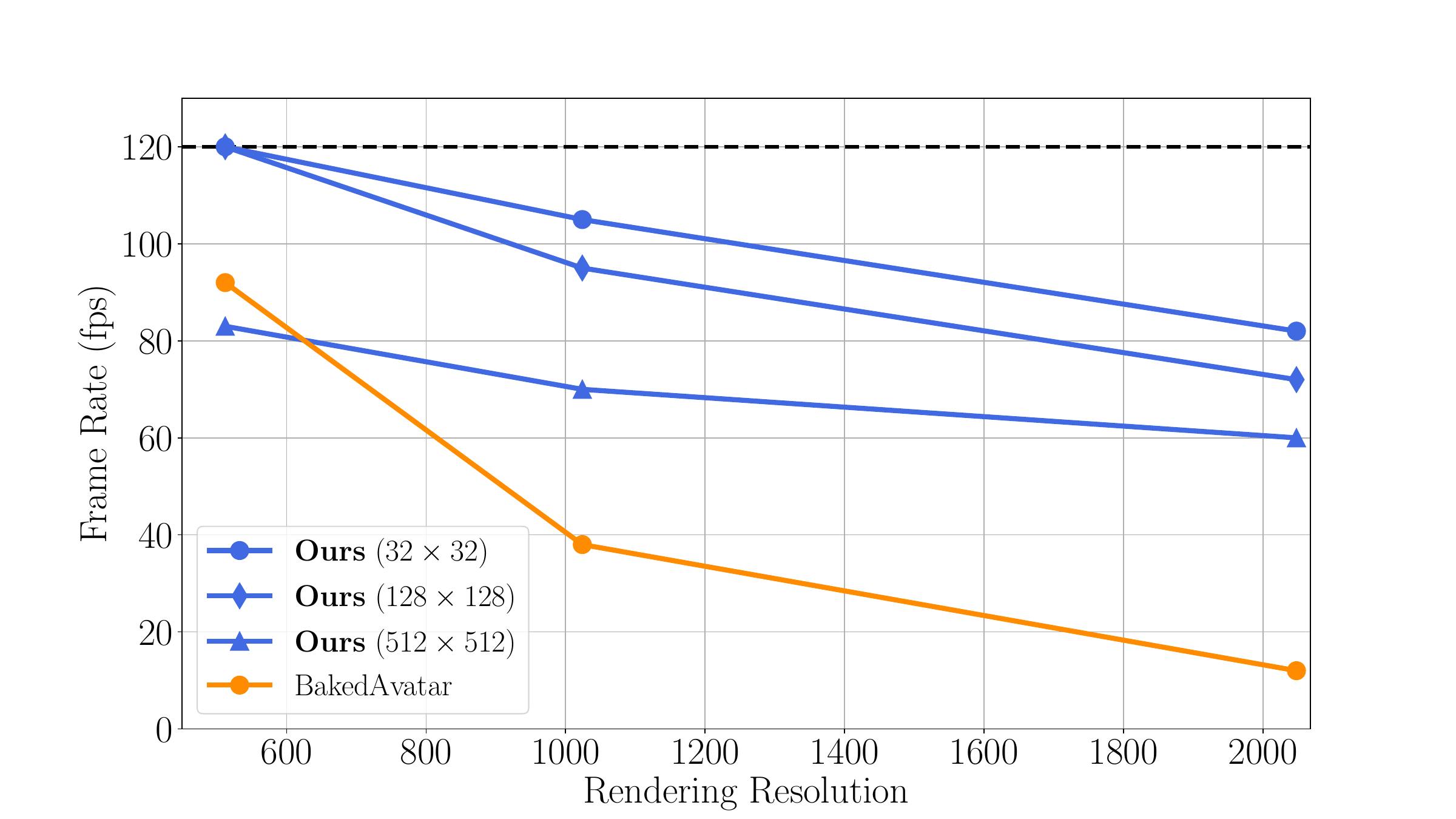}
\caption{\textbf{Frame rates on WebGL.} Our assets can be deployed to traditional graphics engines on web browsers using WebGL, achieving cross-platform compatibility. These numbers are profiled on Chrome 133.0 on a MacBook Pro with M1 Pro chip. Frame rates above the refresh rate of $120$ Hz are indicated as $120$. 
 \vspace{-0.7cm}}
  \label{fig:fps}
\end{figure}

Our representation is natively deployable on graphics platforms and enables real-time rendering of volumetric face avatars using a simple programmable shader. Using WebGL on a consumer laptop, we achieve the frame rates shown in~\cref{fig:fps} for varying mesh resolutions and rendering resolutions, while keeping the memory usage less than $2$ GB at $512^2$ mesh resolution and $2$K rendering resolution.
{We emphasize that our approach discretizes all scene components, placing it on the memory-intensive side of the inherent memory--compute trade-off. Nevertheless, by employing a moderate number of layers and basis sizes, our assets remain sufficiently lightweight to maintain a memory footprint compatible with commodity hardware.}
We also observe that the performance of BakedAvatar~\cite{bakedavatar} suffers significantly at higher rendering resolutions owing to per-pixel MLP queries, while our representation naturally scales well to higher resolutions due to simple texture queries. 

\section{Conclusion}

We present an efficient and natively-deployable representation for animatable volumetric face avatars for traditional graphics pipelines. We achieve this by modeling the canonical face geometry with a static layered mesh, and appearance and deformation as textures. Our model enables efficient control of facial expressions through simple linear blending of our texture assets based on the tracked face parameters. Thorough experimentation and analysis against modern techniques demonstrate the efficacy of our method.

{
    \small
    \bibliographystyle{ieeenat_fullname}
    \bibliography{main}
}
\clearpage
\maketitlesupplementary

\section{Implementation Details}

\noindent\textbf{Architecture details.} The warp and texture predictors are implemented as 6-layer MLPs, where the subject-specific embeddings $\phi_w, \phi_t$ are concatenated to the input coordinates and each of the learned latent codes in embedding matrices $E_w, E_t$ are concatenated to the features after the third layer and fed through the model in parallel to produce a basis of warps and textures during each forward pass. In this formulation, variations across the elements of the bases are offloaded to single matrix, while the weights of the network are shared. This provides sufficient variability within the bases that can generate deformations and appearances with high expressivity, while also maintaining computational efficiency at training time.

\vspace{0.2cm}\noindent\textbf{3DMM and fitting.} 
Our 3DMM includes linear bases of identity and expression. For each frame, we fit the 3DMM by estimating $599$ probabilistic landmarks in 2D and optimizing identity, expression, rotation, and translation parameters of the 3DMM using a loss function that encourages consistency between per-vertex landmarks of the 3DMM and the observed 2D landmarks~\cite{wood20223d}. The parameter size of our expression model is $p=63$.

\begin{figure}[t]
  \centering
  \includegraphics[width=1.0\linewidth]{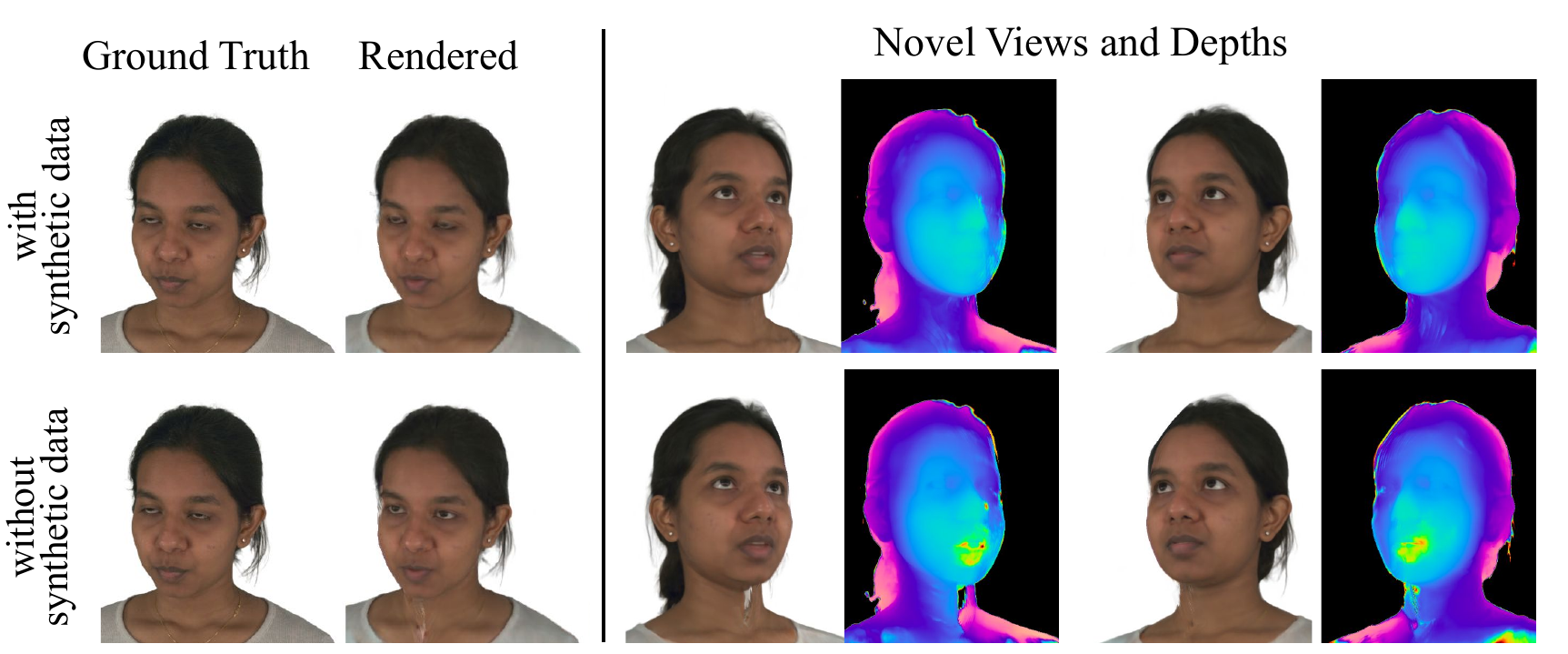}
  \caption{\textbf{Ablation on synthetic data.} The stability and overfitting challenges can be mitigated by introducing synthetic face data jointly trained with the real subject. This helps with generalization to novel expressions in addition to regularization of the learned face geometry.}
  \label{fig:synth-ablation}
\end{figure}

\vspace{0.2cm}\noindent\textbf{Synthetic data.}
We use the synthetic face dataset introduced in~\cite{buehler2024cafca}, where we augment a subset of the subjects in this dataset to the real data. 
During training, we construct our batches by gathering $50\%$ of the rays from the real subject and the other $50\%$ from synthetic subjects. 
To ensure broad coverage of facial dynamics, synthetic expressions are drawn uniformly across the entire dataset.
We illustrate the effectiveness of our joint real--synthetic training in \cref{fig:synth-ablation}, where we observe that in the absence of synthetic data, our model is prone to geometric instabilities and may fail to generalize to novel expressions.

\vspace{0.2cm}\noindent\textbf{Exported assets.} 
The blend weights for our warp and texture bases can be efficiently computed at rendering time by linearly mapping $p=63$ dimensional expression coefficients to $W=T=12$ coefficients. Including the learned constant offset in this mapping, this results in two weight matrices of size $12\times64$.

\begin{figure}[t]
  \centering
  \includegraphics[width=1.0\linewidth]{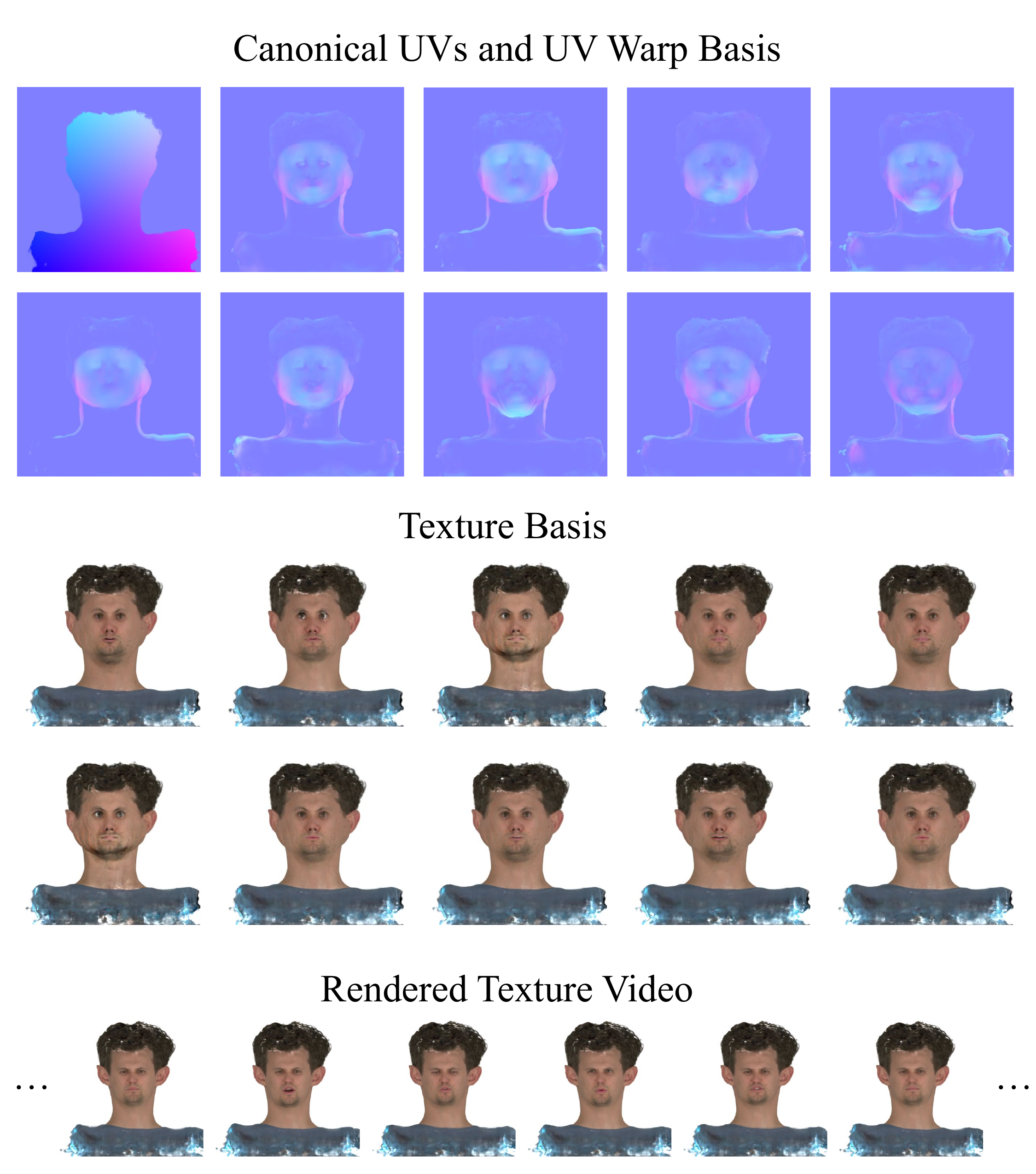}
  \caption{\textbf{Visualizations of the assets.} Illustrating a subset of the learned warps and appearances. With tracked expression coefficients of a 3DMM, these assets can be used to render a texture video shown at the bottom. All images are alpha-composited for visualization purposes.
  \vspace{-0.3cm}}
  \label{fig:asset-viz}
\end{figure}

Our canonical UV values, warp basis, and texture basis are all exported in the UV space, where the resolution and the precision can be modified for different application needs. Please refer to \cref{fig:asset-viz} for visualizations of our assets. For renders at $0.5$K resolution, we found that mesh, warp map, and texture map resolutions of $512\times512$ are sufficient to preserve the overall visual quality. Here, a $32$-bit precision is maintained for UV values, while the appearance is exported as $8$-bit RGBA maps, where the view-dependent radiances are discarded.

\section{Additional Results and Comparisons}

\begin{figure}[t]
  \centering
  \includegraphics[width=1.0\linewidth]{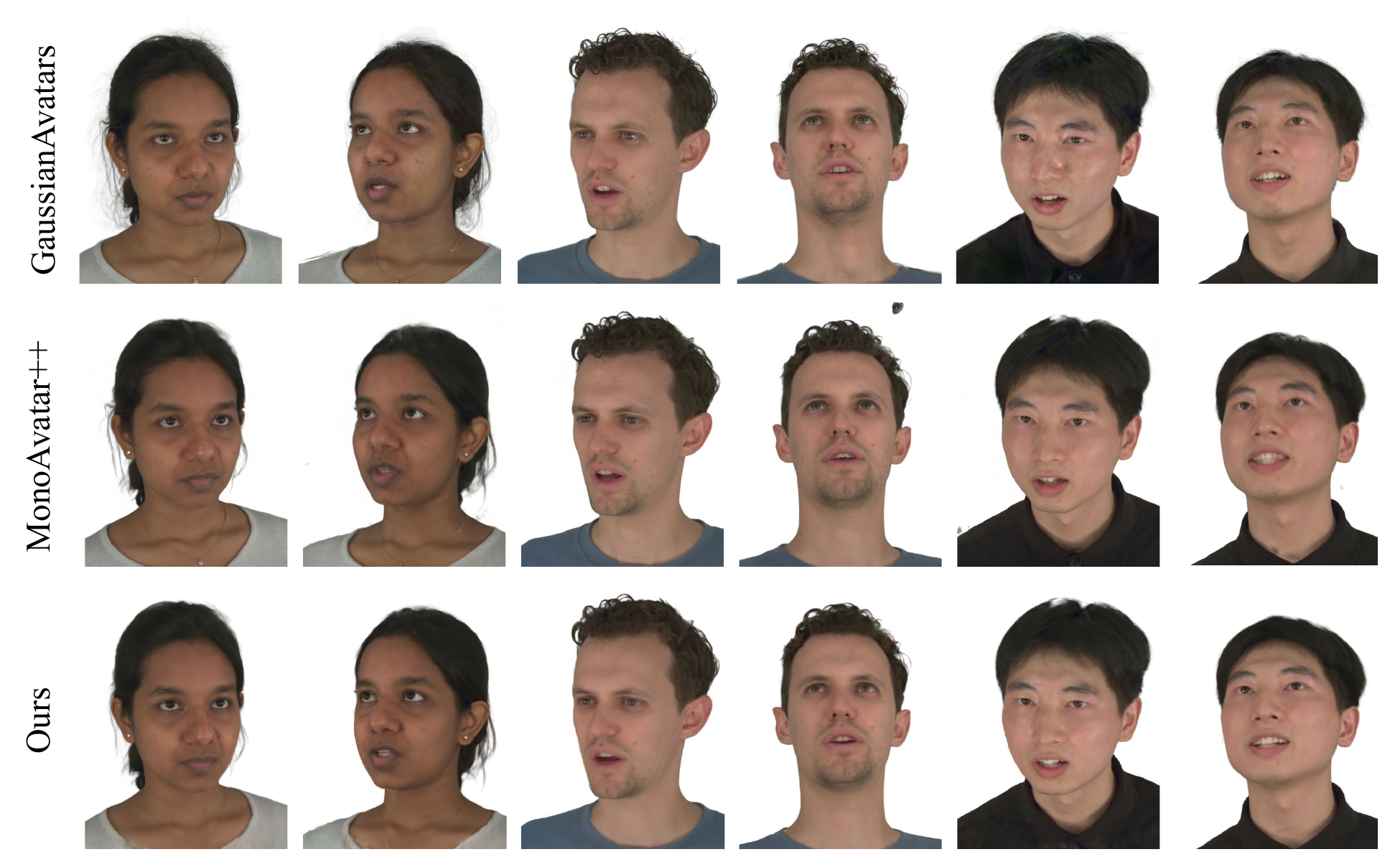}
  \caption{\textbf{Comparisons on novel view synthesis.} Our model can synthesize novel views at a comparable visual quality to  MonoAvatar++~\cite{bai2024efficient} and GaussianAvatars~\cite{qian2024gaussianavatars}, while being less prone to floater artifacts in NeRFs, and inherently preventing primitive ordering artifacts in 3DGS-based methods.}
  \label{fig:orbit-comparison}
\end{figure}

\noindent\textbf{Novel view synthesis comparisons.} We provide comparisons on novel view synthesis with the state of the art methods, see~\cref{fig:orbit-comparison}. NeRF-based methods like MonoAvatar++~\cite{bai2024efficient} can manifest floating artifacts and 3DGS-based methods may result in popping-like artifacts due to explicit sorting of primitives~\cite{radl2024stopthepop}. Our method is not prone to such artifacts by design, and the exported textured meshes can be edited by an artist, providing additional flexibility to remove visual seams as a post-processing step. Please see the supplementary video for better visualizations.

\begin{figure}[t]
  \centering
  \includegraphics[width=1.0\linewidth]{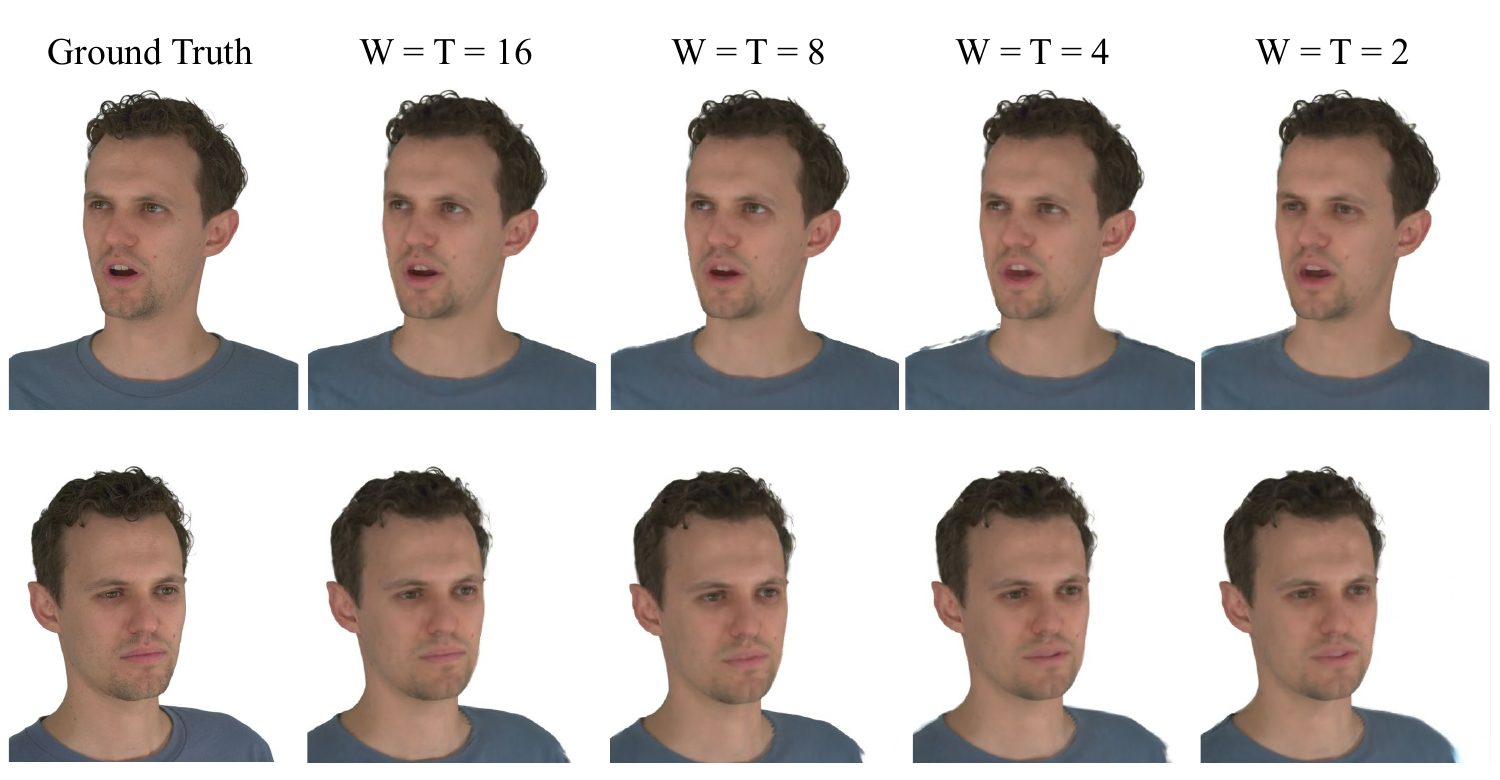}
  \caption{\textbf{Ablation on sizes of warp and texture bases.} Sufficient number of blendable warps and textures is crucial to achieve good rendering quality and generalization to novel expressions.\vspace{-0.3cm}}
  \label{fig:ablation}
\end{figure}

\vspace{0.2cm}\noindent\textbf{Warp and texture basis size ablations.}
To provide more insights into our model, we evaluate its expressiveness by modifying its warp and texture basis sizes. We illustrate our results in~\cref{fig:ablation} and evaluation metrics in~\cref{table:ablation}, where we observe that the overall rendering quality suffers and the renders manifest artifacts as we reduce the basis sizes. Furthermore, the model does not generalize to novel facial expressions and eye gazes as we reduce its capacity.

\begin{table}
\footnotesize
  \caption{\textbf{Ablation study on sizes of warp and texture bases.} Texture and warp basis sizes improve rendering quality and generalization. These metrics are obtained on cropped images that include the face region only.}
  \label{table:ablation}
  \begin{tabular}{cccc}
    \toprule
     & PSNR\,$\uparrow$ & SSIM\,$\uparrow$ & LPIPS\,$\downarrow$ \\
    \midrule
    $W=T=16$ & $29.67 \pm 1.36$ &  $0.897 \pm 0.012$ & $0.258 \pm 0.017$ \\
    $W=T=8$ & $29.47 \pm 1.39$ & $0.894 \pm 0.012$ & $0.259 \pm 0.017$\\
    $W=T=4$ & $29.38 \pm 1.41$ & $0.892  \pm 0.012$ & $0.263 \pm 0.016$\\
    $W=T=2$ & $28.46 \pm 1.21$ & $0.882  \pm 0.012$ & $0.276 \pm 0.019$\\
  \bottomrule
\end{tabular}
\end{table}

\section{Limitations and Future Work}

\begin{figure}[t]
  \centering
  \includegraphics[width=0.9\linewidth]{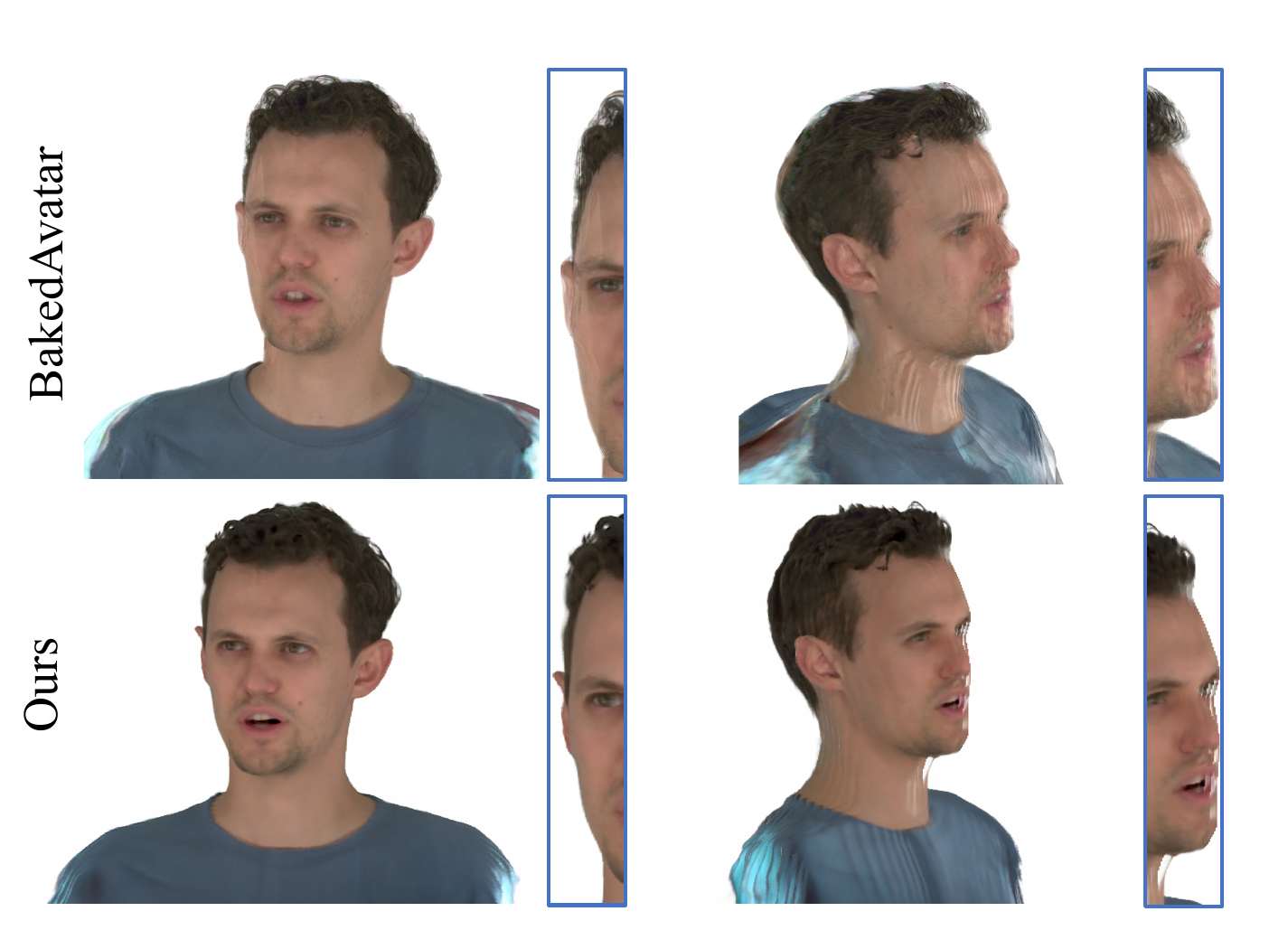}
  \caption{\textbf{Limitations}. Layered mesh representations may suffer from \emph{shell} artifacts at extreme angles. While our approach outperforms existing baselines on visual quality on novel views at moderate poses \textit{(left)}, at more extreme out-of-training profile views \textit{(right)} it also suffers from \emph{shell} artifacts similar to the baseline.}
  \label{fig:limitations}
\end{figure}

The basic premise of our representation is the discretization of scene components like geometry, appearance and deformations. While this enables traditional rendering, it also inherently limits the representational capacity compared to continuous volumes such as radiance fields and 3D Gaussians. 
Furthermore, since we project expression parameters onto low-dimensional blend weights, our model may exhibit blurring artifacts for extreme expressions, particularly those involving strong deformations around the mouth.
We also note that our model does not explicitly account for the head pose, and hence the neck and torso regions may show slight instabilities in cases where the training sequences involve strong head pose variations.

Our layered mesh with transparency can be seen as a generalization of multiplane imaging (MPI)~\cite{zhou2018stereo}, where we instead learn a set of surfaces that follow a coarse face geometry and represent dynamic scenes. Due to such coarse geometry, there exists a fundamental limit to the viewing angle range for artifact-free rendering~\cite{srinivasan2019pushing}. At extreme angles, our representation can manifest shell artifacts, please see~\cref{fig:limitations}. {We have also not tested our representation for controlling/animating large deformations such as head/neck rotations. These may also require additionally including and optimizing for a root joint UV-deformation of the layered mesh in the neck region.} Optimizing and regularizing the topology of the layered mesh and the texture basis to improve the overall representational capacity could be an interesting future line of research.

The enrollment phase in our pipeline relies on ML training and inference, future work could explore simplifying this process. Recently, generative models have been used to learn a strong prior of face geometry and appearance \cite{live_3d_portrait}, allowing direct regression of the face volume from even single images. Such quick and efficient techniques can further help reduce the compute and memory cost of the enrollment phase by learning a data-driven generative model. This can particularly help extend our representation to consumer use-cases such as monocular enrollment and tracking.

We believe that our work lays the important groundwork of building a novel representation that is capable of representing, animating and synthesizing volumetric effects in traditional graphics pipelines, and future work can build on and expand it towards even more practical settings.

\end{document}